\documentclass[journal]{IEEEtran}

\ifCLASSINFOpdf
\usepackage[pdftex]{graphicx}
\graphicspath{{./}}

\DeclareGraphicsExtensions{.png,.pdf,.jpg,.jpeg}
\else
   \usepackage[dvips]{graphicx}
   \graphicspath{{Figures/}}
     \DeclareGraphicsExtensions{.png,.pdf,.jpg,.jpeg}
\fi

\usepackage{arydshln} 
\usepackage{fancyhdr}  

\usepackage{algorithmic}
\usepackage{algorithm}
\usepackage{array}
\usepackage[caption=false,font=normalsize,labelfont=sf,textfont=sf]{subfig}
\usepackage{textcomp}
\usepackage{stfloats}
\usepackage{url}
\usepackage{verbatim}
\usepackage{cite}
\usepackage{multirow}
\usepackage{pifont}
\usepackage{amsmath,amssymb,amsfonts,amsbsy}
\usepackage{bm} 
\usepackage{graphicx}
\usepackage{mathtools}
\usepackage{array,ragged2e}
\usepackage{booktabs}
\usepackage{makecell}
\usepackage{placeins}

\usepackage{color}
\usepackage[table]{xcolor}
\definecolor{myPink}{rgb}{0.9294, 0.0078, 0.5490}
\definecolor{Gray}{gray}{0.92}
\definecolor{my_color}{HTML}{E8F3F1}
\definecolor{lightgray}{gray}{0.9}  
\usepackage[hidelinks]{hyperref}
\hypersetup{
  pdftitle={UniPCB: A Generation-Assisted Vision-Based Measurement Framework for PCB Defect Inspection},
  pdfauthor={Huan Zhang et al.},
  pdfkeywords={automated optical inspection, PCB defect detection, vision-based measurement}
}
\usepackage{threeparttable}
\usepackage{diagbox}
\usepackage{orcidlink}

\newcommand{\rev}[1]{#1}

\makeatletter
\let\NAT@parse\undefined
\makeatother

\begin{document}

\title{\rev{UniPCB: A Generation-Assisted Vision-Based Measurement Framework for PCB Defect Inspection}} 




\author{Huan~Zhang,
        Lianghong Tan,
        Yichu Xu, Zishan Su,
        Jiangzhong Cao$^{\dagger}$,
        Huanqi~Wu, 
        Linwei Zhu,
        and
        Xu~Zhang$^{\dagger}$


\thanks{This work was in part supported by the National Natural Science Foundation of China under Grant 62302105, in part by the Guangdong Provincial 
Key Laboratory of Intellectual Property \& Big Data under Grant 
2018B030322016. \emph{(Corresponding author: Jiangzhong Cao, Xu Zhang.)}}%

\thanks{Huan Zhang, Lianghong Tan, Jiangzhong Cao, and Huanqi Wu are with the  School of Information Engineering, Guangdong University of Technology, Guangzhou 510006, China (e-mail: huanzhang2021@gdut.edu.cn; 656896486@qq.com; cjz510@gdut.edu.cn; 2441025664@qq.com).}

\thanks{Yichu Xu, Zishan Su, and Xu Zhang are with the School of Computer Science, Wuhan University, Wuhan 430072, China (e-mail: xuyichu@whu.edu.cn; zishan.su@whu.edu.cn; zhangx0802@whu.edu.cn).}

\thanks{Linwei Zhu is with the Shenzhen Institutes of 
Advanced Technology, Chinese Academy of Sciences, Shenzhen 518055, China (e-mail: lw.zhu@siat.ac.cn).}
}


\markboth{\rev{IEEE Transactions on Instrumentation and Measurement}}%
{\rev{Zhang \MakeLowercase{\textit{et al.}}: UniPCB for PCB Defect Inspection}}

\maketitle

\begin{abstract} %
\rev{Automated optical inspection (AOI) is a vision-based instrumentation process that detects and localizes physical defects on printed circuit boards (PCBs) from optically acquired images. Its reliability is often limited by scarce and class-imbalanced defect observations and by insufficient representation of small, low-contrast defects against dense circuit patterns. We propose UniPCB, a generation-assisted PCB inspection framework that combines controlled defect synthesis with task-specific detection. The generation branch derives heterogeneous edge, relative pseudo-depth, and text conditions from PCB images. A ScaleEncoder embeds the conditions at four U-Net resolutions, while a Condition Modulation block performs FiLM-style spatially adaptive fusion. The detection branch introduces an Inverted Residual Shift Attention block for joint global--local modeling and a Cross-level Complementary Fusion block for selective hierarchical feature integration. The generated images augment the detector training data and improve defect coverage without changing the online inspection path. Experiments using the corrected annotations of DsPCBSD+ yield an mAP@0.5 of 98.0\% and an mAP@0.5:0.95 of 61.8\%. Under the same generation-augmented training setting, these values exceed those of the RT-DETR baseline by 2.1 and 1.7 percentage points, respectively. The generation branch obtains an FID of 129.61 and an SSIM of 0.619. These results indicate the potential of structured synthetic data and PCB-oriented feature modeling for improving vision-based defect inspection. Code is available at \url{https://github.com/House-yuyu/UniPCB}.}

\end{abstract}

\IEEEpeerreviewmaketitle

\section{Introduction}
\label{sec:intro} 

{
\IEEEPARstart{P}{rinted} circuit boards (PCBs) are essential components in modern electronic products, and their manufacturing quality directly affects the reliability of downstream devices. Complex fabrication processes may introduce physical defects such as shorts, opens, mouse bites, spurs, and spurious copper, which can degrade electrical connectivity and cause product failure. Automated optical inspection (AOI) can be formulated as a vision-based instrumentation process in which the camera, illumination, image-acquisition chain, and computational model jointly detect and localize these defects~\cite{VBM,appliedAI}. The resulting defect class, confidence, and bounding-box location constitute the inspection output used for manufacturing quality decisions. Consequently, missed detections, false alarms, and localization errors directly affect the reliability of the inspection process. Traditional reference-image comparison and handcrafted feature matching are sensitive to lighting variations, board-layout changes, and novel defect morphologies, often degrading these outputs in practical inspection scenarios~\cite{dd1}.
\par}

Deep learning-based detectors improve robustness by learning discriminative representations directly from data. However, their effectiveness depends on large and well-annotated defect datasets, which are difficult to obtain in PCB manufacturing. First, mature production lines usually have high yield rates, making defective samples naturally rare. Second, PCB layouts vary across products, limiting the transferability of defect samples across board types. Third, defect annotation requires domain expertise and is time-consuming, especially when dense and small defects need to be precisely localized. As a result, PCB defect datasets are often scarce and class-imbalanced, causing detectors to underfit rare defects and generalize poorly to unseen board layouts.

To alleviate this data bottleneck, traditional augmentation techniques such as rotation, flipping, and blurring have been widely adopted. Nevertheless, these operations mainly transform existing samples and cannot sufficiently model the diverse defect morphologies and local structural variations observed in real PCB images. Deep generative models provide a more flexible alternative. VAEs~\rev{\cite{vae}} and GANs~\cite{gan} have been explored for defect synthesis, but VAEs often produce blurry details and GANs may suffer from training instability and mode collapse. Diffusion models~\rev{\cite{ddpm,ddim,LDM,Any2RSI}} offer stronger training stability and better texture modeling ability, making them suitable for fine-grained industrial defect synthesis. However, 
their training on natural image datasets limits their effectiveness 
on industrial images such as PCBs~\rev{\cite{croitoru2023diffusion}}, and existing 
PCB-oriented methods typically rely on a single control condition 
injected at the initial resolution, providing insufficient structural 
guidance for the dense, fine-grained circuit patterns unique to PCB 
images.

\begin{figure}[!tb]  
	\centerline{\includegraphics[page=1,trim = 0mm 0mm 0mm 0mm, clip, width=1\linewidth]{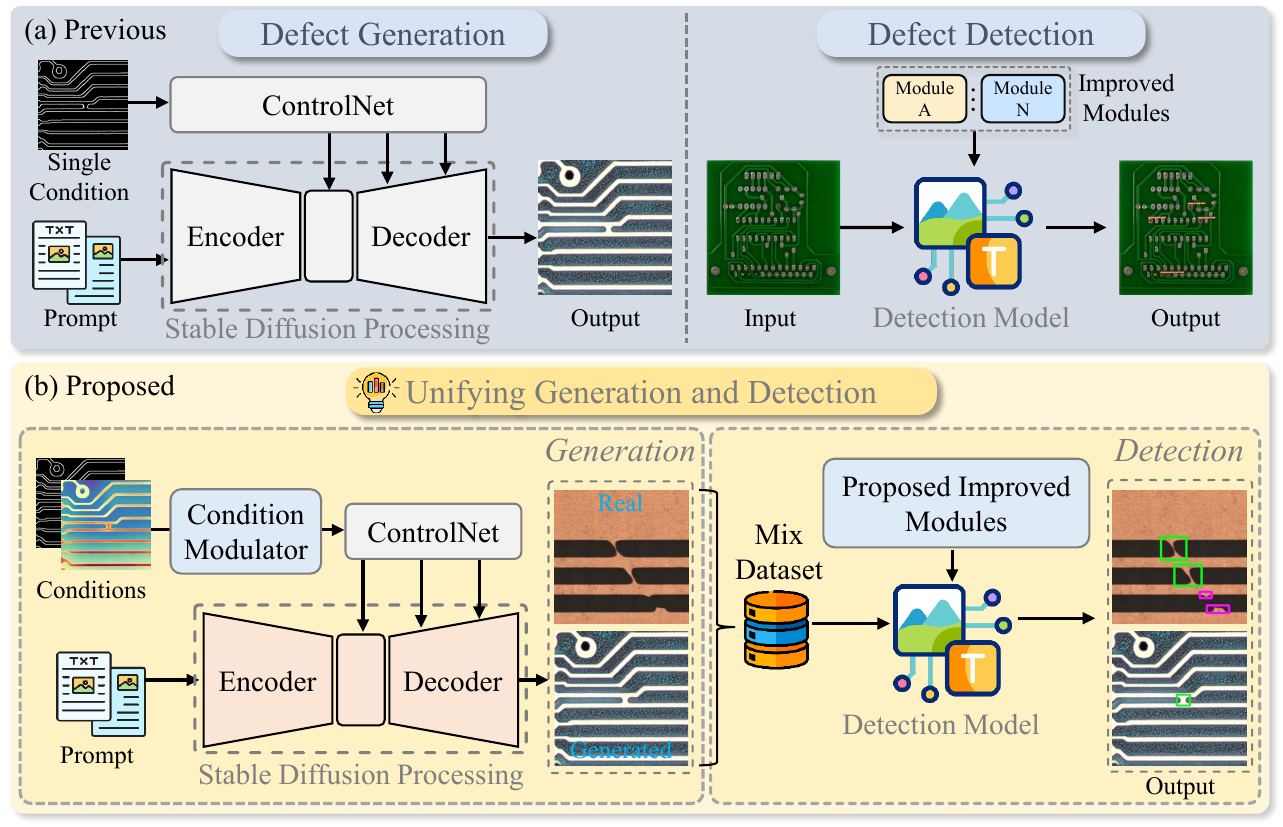}}
  \captionsetup{skip=0pt}
	\caption{
Comparison between UniPCB and previous approaches 
for PCB defect detection. (a) Previous methods address the two 
challenges in isolation: generation methods rely on single-modality 
conditions with limited structural control, while detection methods 
improve architectures but remain bottlenecked by scarce training data. 
(b) UniPCB jointly addresses both within a unified framework: the 
generation branch synthesizes high-quality defect samples while the detection branch exploits the enriched data 
through task-specific feature modeling modules.
    }  
	\label{fig:motivate}
	\vspace{-1em}
\end{figure}

In addition to data scarcity, the feature representation capability of the detector remains critical. Transformer-based detectors such as DETR~\cite{DETR} and RT-DETR~\cite{RT-DETR} provide strong global modeling ability. \rev{Lightweight detection designs likewise seek to balance accuracy and computational cost.} Recent studies in surface defect and tiny object detection further indicate that fine-grained feature separation, relation reasoning, class knowledge, and regional attention are important for detecting subtle defects under complex backgrounds~\cite{shen2026class, unfolddet, tiny_rcsa}. Nevertheless, architecture-level improvements alone may still be constrained by insufficient and imbalanced training data \cite{dd4}.

These two issues often appear together in PCB inspection and jointly limit practical detection performance. On the one hand, a detection model trained on scarce 
and imbalanced data cannot fully leverage improvements in model 
architecture, as the data bottleneck fundamentally limits what the 
model can learn. On the other hand, a generation model producing 
low-fidelity or poorly controlled defect samples provides little 
benefit to detection, even when the detector itself is powerful. 
This coupling implies that addressing either challenge in isolation 
yields diminishing returns: the two must be resolved jointly within 
a coherent framework to achieve meaningful gains in PCB defect 
inspection.

To break this mutual bottleneck, we propose a unified 
generative--detection framework, termed \textbf{UniPCB}, which 
simultaneously enhances data quality through controlled defect 
synthesis and improves detection capability through task-specific 
architectural design. Fig.~\ref{fig:motivate} compares UniPCB with 
previous approaches. As illustrated in Fig.~\ref{fig:motivate}(a), 
prior work treats the two problems separately: diffusion-based methods 
reconstruct images to identify pixel-level discrepancies but cannot 
generate new training data, while object detection methods improve 
architectures but remain fundamentally bottlenecked by scarce labeled 
data. As shown in Fig.~\ref{fig:motivate}(b), UniPCB unifies both 
within a single framework. On the generation side, a multi-modal 
conditional strategy synthesizes diverse and realistic defect samples 
by injecting edge, depth, and text conditions at multiple scales, 
ensuring structural alignment and semantic fidelity to real PCB 
images. On the detection side, the generated data directly serves as augmented 
training input. To further strengthen the detector's sensitivity to 
small, low-contrast defects under complex circuit backgrounds, two 
task-specific modules are introduced: the IRSA Block for joint 
global--local feature modeling, and the CLCF Block for cross-level 
selective feature fusion. Through this co-design of generation and 
detection, each component amplifies the benefit of the other.

In summary, our main contributions are as follows.  

\noindent \ding{113}~
\rev{We propose UniPCB as a generation-assisted vision-based measurement framework for PCB defect inspection. It couples controlled defect synthesis with a task-specific detector and evaluates how structured synthetic samples and PCB-oriented feature modeling improve the reliability of defect classification and localization.}

\noindent \ding{113}~
\rev{On the generation side, we design a Multi-modal Condition Generator that constructs parallel edge, relative pseudo-depth, and text conditions, providing heterogeneous and structured control signals to alleviate data scarcity and class imbalance. Building on this, we propose a latent diffusion-based synthesis network with a ScaleEncoder that embeds conditions at four resolutions and a CondMod block that jointly modulates noise and textual features, improving sample fidelity and structural consistency.}

\noindent \ding{113}~
\rev{On the detection side, we propose an Inverted Residual Shift Attention Block that couples self-attention and shift-wise convolution in cascade within an inverted-residual structure, and a Cross-level Complementary Fusion Block that selectively integrates fine spatial detail with high-level semantics. Together, the two modules improve the classification and localization of small, low-contrast PCB defects.}

\section{Related Work} 
\subsection{\rev{Vision-Based PCB Inspection as an I\&M System}}

\rev{Vision-based measurement uses an imaging sensor and computation to detect, monitor, or quantify a physical phenomenon~\cite{VBM}. In PCB AOI, the phenomena are manufacturing defects, and the reported quantities include their presence, category, confidence, and image-plane location. UniPCB operates in the processing and analysis stage of this measurement chain; its performance is therefore tied to the acquisition conditions, data distribution, and reliability of the inspection decisions~\cite{appliedAI}.}

\subsection{PCB Defect Generation}

In PCB defect detection, deep learning models require sufficient labeled data, yet real defect samples are scarce, imbalanced, and costly to annotate. Synthetic defect generation has therefore been explored as an effective way to enrich training data and improve detector generalization. Existing methods mainly follow two technical routes: GAN-based synthesis and diffusion-based conditional generation.

Early defect generation methods often adopted GAN-based frameworks to synthesize additional defective samples for data augmentation~\cite{contrastive, duan2023fewshot, sd-yolov5}. By learning the distribution of defect textures and backgrounds, these methods can increase sample diversity under limited-data settings. Related conditional generation techniques, such as self-attention based generation~\cite{zhang2019sagan} and spatially-adaptive normalization~\cite{park2019spade}, further show the importance of long-range dependency modeling and spatially-aware modulation for preserving structural consistency. However, GAN-based methods are generally limited by unstable training, mode collapse, and insufficient control over the spatial relationship between defects and complex PCB backgrounds.

Recently, diffusion models have become a more promising alternative because of their stable training process, strong texture modeling ability, and flexible conditional control~\cite{trabucco2024effective,hu2024anomalydiffusion}. In PCB-related defect generation, Deng et al.~\cite{dm1} introduced feature modulation and normalized flow modules to coordinate local and non-local information for learning high-quality defect distributions. Beyond PCB-specific settings, controllable diffusion methods such as T2I-Adapter~\cite{mou2024t2iadapter} and Uni-ControlNet~\cite{qin2023unicontrol} demonstrate that external structural conditions can effectively guide pretrained diffusion models and improve controllability. These studies suggest that diffusion-based generation is well suited for synthesizing fine-grained industrial defect images.

Despite these advances, existing defect generation methods still have two limitations. First, most methods rely on limited or single-form conditions, which are insufficient to describe the dense circuit structures and subtle defect patterns in PCB images. Second, conditional signals are often injected in a coarse or single-scale manner, making it difficult to preserve fine traces and local defect geometry across the denoising process. To address these issues, we introduce a multi-modal condition generator together with multi-scale condition embedding and modulation, enabling more structured and defect-aware PCB sample synthesis for downstream detection.

\subsection{PCB Defect Detection} 


PCB defect detection differs from natural image detection because PCB defects are usually small, sparse, irregularly shaped, and visually similar to normal circuit patterns. Traditional handcrafted or template-matching methods are sensitive to manufacturing variations and therefore generalize poorly across board layouts and imaging conditions. Deep learning based detectors have improved robustness by learning discriminative representations automatically, and recent PCB defect detection studies~\cite{yolo-hmc, LAC, ADA, wang2023internimage, sgt-yolo} mainly improve performance from two perspectives: adaptive feature modeling and multi-scale feature fusion. Beyond PCB-specific scenarios, recent TCSVT studies on industrial defect and anomaly detection have also explored knowledge distillation for rail surface defects~\cite{modal_eval_kd_zhou2024}, differential distillation for unsupervised anomaly localization~\cite{pull_push_zhou2023}, multi-perspective multimodal fusion for 3D industrial defects~\cite{m3df_asad2025}, and memory-augmented wavelet representations for anomaly detection~\cite{sam_wcnn_wu2023}. These works further indicate that efficient representation learning and robust localization are central issues in industrial visual inspection.

Adaptive feature modeling \cite{fe1, fe3} aims to enhance defect-relevant responses while suppressing complex circuit background interference. Liu et al.~\cite{fe1} introduced a multi-head nonlocal transformer module to capture long-range dependencies and focus on defect regions. Guo et al.~\cite{fe3} proposed an Image-to-Instance Contextual Enhancement Component that incorporates global contextual information to reweight proposal features. Dai et al.~\cite{dai2021dynamichead} designed Dynamic Head, which dynamically modulates detection features with scale-aware, spatial-aware, and task-aware attention. These methods demonstrate the importance of adaptive representation learning for subtle defect perception.

Another line of work focuses on multi-scale feature fusion \cite{ff1, mrc, ff3}, which is critical for localizing tiny defects while preserving semantic discrimination. Yang et al.~\cite{ff1} designed a Coordinate Feature Refinement module to fuse channel- and spatial-level features and improve scale robustness. Cao et al.~\cite{mrc} proposed MRC-DETR, which embeds multi-scale residual coupling into a DETR-based framework for bare-board PCB defect detection. Ji et al.~\cite{ff3} introduced a Slim-Scale Adaptive Fusion structure that aggregates hierarchical features using grouped spatial convolutional pyramids and channel shuffle, improving fusion efficiency and detection accuracy.

Although these PCB-specific and general industrial inspection methods improve feature representation, localization, and deployment efficiency, they mainly operate at the detector or anomaly modeling level and typically assume that the training distribution is sufficiently representative. In practical PCB inspection, however, defect samples remain scarce and highly imbalanced, especially for rare defect categories. As a result, architecture-level improvements alone may still be limited by insufficient training data. This motivates our generation-assisted detection pipeline, which combines structured defect synthesis with task-specific feature modeling to address both data scarcity and representation limitations.

\begin{figure*}[!t]  
	\centerline{\includegraphics[page=1,trim = 0mm 0mm 0mm 0mm, clip, width=0.9\linewidth]{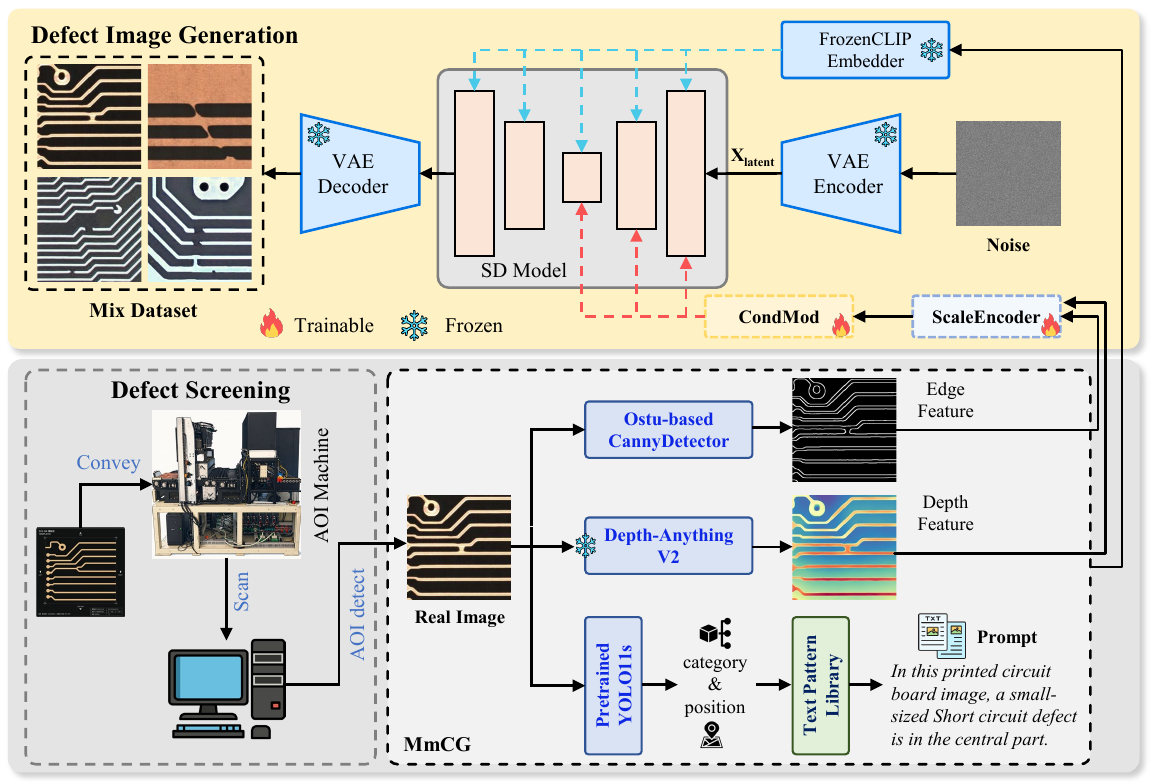}}
  \captionsetup{skip=0pt}
	\caption{\rev{Overview of the proposed defect generation framework. DsPCBSD+ images are processed by the Multi-modal Condition Generator (MmCG) to extract edge, relative pseudo-depth, and text conditions. ScaleEncoder and CondMod embed these conditions into Stable Diffusion, and the synthesized samples augment the detector training data. The AOI block is a workflow schematic, not a private data source.}}  
	\label{fig:g_fra}
	\vspace{-1em}
\end{figure*}

\section{Method}
\label{Method}
\subsection{Overview architecture}

\subsubsection{\textbf{Generation Framework}}
As illustrated in Fig.~\ref{fig:g_fra}, the generation network is 
built upon Stable Diffusion~\cite{LDM} with ControlNet~\cite{controlnet} 
and extended with three dedicated modules for non-natural industrial 
images. A real PCB image first passes through the Multi-modal Condition 
Generator (MmCG), which produces complementary edge, \rev{relative pseudo-depth}, and text 
conditions in parallel. These conditions are then embedded into the 
diffusion U-Net at four resolutions by the ScaleEncoder, and at each 
resolution a CondMod block jointly modulates the noise feature with 
  the condition and textual embedding to steer the denoising process. 
  \rev{The synthesized defect images are merged with real data to form a 
  generation-augmented training set with improved defect coverage for downstream detection.} The three modules are 
detailed in Sec.~\ref{sec:gen}.

\subsubsection{\textbf{Detection Framework}}
\begin{figure*}[!t]  
	\centerline{\includegraphics[page=1,trim = 0mm 0mm 0mm 0mm, clip, width=0.85\linewidth]{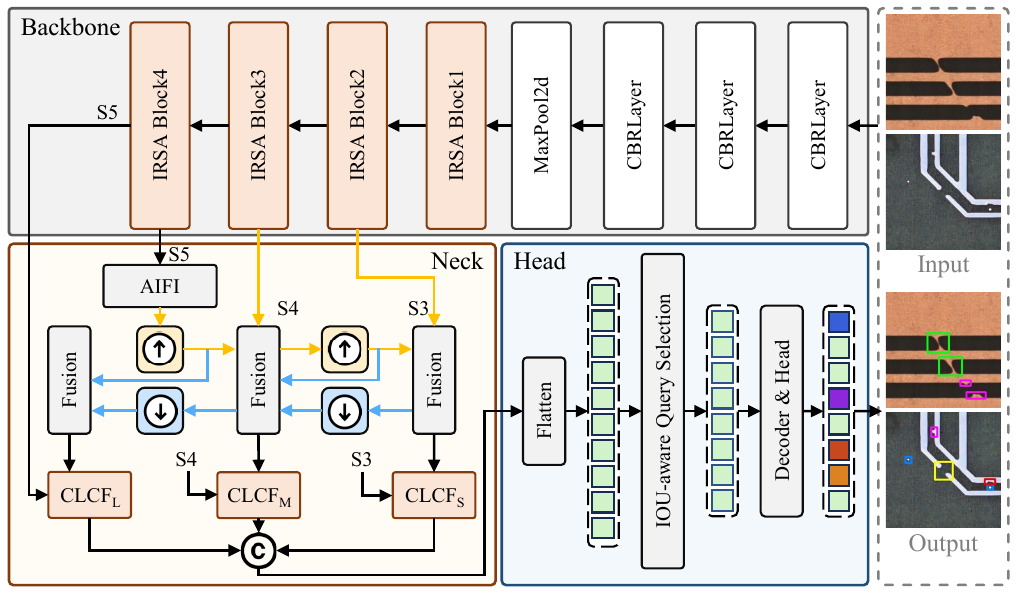}}
  \captionsetup{skip=0pt}
	\caption{Overview of the proposed detection framework. IRSA Blocks replace BasicBlocks in all four stages of the ResNet-18 backbone. Yellow and blue blocks in the Neck denote upsampling and downsampling operations, respectively.}  
	\label{fig:d_fra}
	\vspace{-1em}
\end{figure*}

As shown in Fig.~\ref{fig:d_fra}, the detection network extends 
RT-DETR~\cite{RT-DETR} with a ResNet-18 backbone and two task-specific 
modules. IRSA Blocks replace the original BasicBlocks at every backbone 
stage to produce more discriminative features, and three CLCF blocks 
(CLCF$_L$, CLCF$_M$, CLCF$_S$) subsequently fuse the backbone features 
with the CCFM outputs at pyramid levels S3, S4, and S5. The fused 
features are concatenated and fed to the IoU-aware query selection and 
decoder. The two modules are detailed in Sec.~\ref{sec:det}.

\begin{figure}[!h]  
	\centerline{\includegraphics[page=1,trim = 0mm 0mm 0mm 0mm, clip, width=1\linewidth]{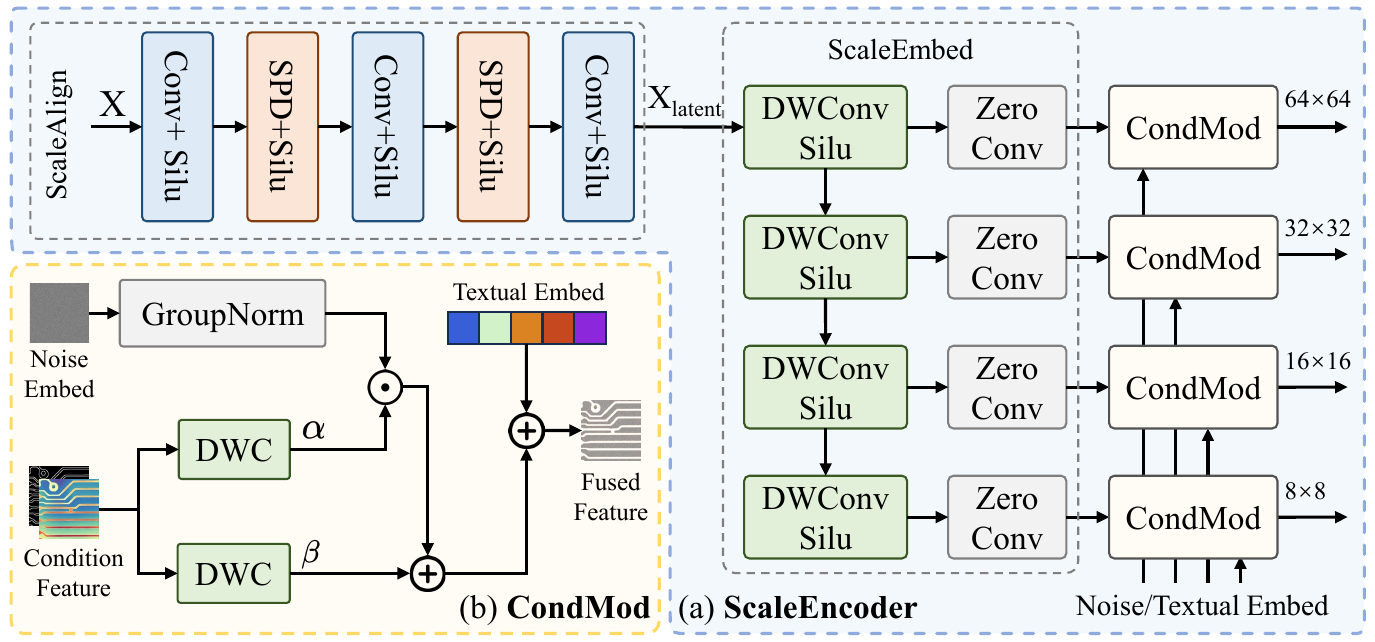}}
  \captionsetup{skip=0pt}
	\caption{(a) Structure of the ScaleEncoder, which encodes the condition map via ScaleAlign layer and injects multi-resolution features into the diffusion U-Net via ScaleEmbed layer. (b) Structure of the Condition Modulation (CondMod), which applies FiLM-style spatially-adaptive modulation to fuse condition features and textual embeddings with the noise map at each resolution.}  
	\label{fig:g_ipr}
	\vspace{-1em}
\end{figure}

\subsection{Generation Framework} 
\label{sec:gen}

\subsubsection{\textbf{Latent Diffusion Model}}
We build our generator on the Latent Diffusion Model (LDM)~\cite{LDM}, 
which performs denoising in the latent space of a pretrained 
autoencoder rather than in pixel space, substantially reducing 
computational cost. Given an input image $x \in \mathbb{R}^{H\times W\times 3}$, 
the encoder $\mathcal{E}$ compresses it into a latent code 
$z = \mathcal{E}(x)$, on which forward diffusion and reverse denoising 
are performed following the standard DDPM 
formulation~\cite{ddpm}:
\begin{equation}
q(z_t \mid z_0) = \mathcal{N}\!\left(z_t;\, \sqrt{\bar{\alpha}_t}\, z_0,\, 
(1-\bar{\alpha}_t)\mathbf{I}\right),
\quad \bar{\alpha}_t = \prod_{i=1}^{t}(1-\beta_i),
\end{equation}
where $\{\beta_t\}_{t=1}^T$ is a fixed variance schedule. At inference, 
the denoised latent $\tilde{z}$ is decoded back to image space as 
$\tilde{x} = \mathcal{D}(\tilde{z})$. We adopt DDIM 
sampling~\cite{ddim} for efficient skip-step generation. 
The condition injection and modulation mechanisms specific to our 
method are detailed in the following subsections.

\subsubsection{\textbf{Multi-modal Condition Generator}}
Diffusion models require paired (image, condition) data, yet no public 
image–text PCB dataset exists. To automate dataset construction and 
provide rich control signals, we design a Multi-modal Condition 
Generator (MmCG) with three parallel branches, illustrated in Fig.~\ref{fig:g_fra}.

\emph{Edge branch.} PCB traces and substrates differ sharply in color 
and brightness, making edge maps a natural structural prior. We employ 
the Canny operator~\cite{canny} with high and low thresholds adaptively 
derived from the Otsu threshold~\cite{otsu} (scaled by fixed upper and 
lower factors), yielding image-specific edges without manual tuning.

\rev{\emph{Relative pseudo-depth branch.} Edge maps alone cannot distinguish foreground 
traces from background substrate, which may cause spatial 
misalignment. We therefore estimate a depth map using the pretrained 
Depth-Anything V2~\cite{depth-anything-v2}. Because monocular depth estimation does not measure the physical height of a planar PCB, we treat its output as a relative pseudo-depth prior that supplies complementary scene-layout cues.}

\rev{\emph{Text branch.} Since no public image--text PCB dataset exists, 
we construct structured prompts automatically from detection outputs. 
A pretrained YOLO11~\cite{yolo11} first predicts defect categories and 
bounding boxes for images in the DsPCBSD+ generation subset; no separately collected private images are used in the reported experiments.} Each defect is then 
characterized along three axes: \emph{scale}, classified as small, 
medium, or large via a dual-threshold rule applied to the bounding-box 
area; \emph{location}, encoded by mapping the normalized box-center 
coordinates onto a $3{\times}3$ spatial grid (top, bottom, left, right, 
center, and four corners); and \emph{distribution}, described at the 
instance level when defect count is low (per-defect category, scale, 
and position) or abstracted to region-level patterns (e.g., scattered 
or locally clustered) when count is high. The resulting descriptions 
are slotted into a template library parameterized by 
\{category, scale, location, quantity\} to produce structured prompts, 
which are subsequently encoded by the frozen CLIP text encoder.

Together, the three branches provide structural, spatial, and semantic 
conditions that jointly guide the downstream diffusion process.

\subsubsection{\textbf{ScaleEncoder}}
ControlNet~\cite{controlnet} injects condition features only at the 
initial resolution, which is inadequate for PCB images with dense, 
fine-grained circuit patterns. Following the multi-scale injection 
principle of Uni-ControlNet~\cite{uni-controlnet}, we propose a 
ScaleEncoder that embeds conditions into the diffusion U-Net at four 
resolutions (64$\times$64, 32$\times$32, 16$\times$16, 8$\times$8).

As shown in Fig.~\ref{fig:g_ipr}(a), ScaleEncoder consists of two 
components. \emph{ScaleAlign} encodes the raw condition map through 
three SiLU-activated standard convolutions interleaved with two 
SPD-Conv downsampling layers~\cite{sunkara2022no}. SPD-Conv reorganizes 
spatial information into the channel dimension, preserving local 
structural cues that standard strided convolutions tend to lose, a 
property well suited to thin PCB traces. \emph{ScaleEmbed} then taps 
intermediate features at four resolutions and refines each via a 
DWConv–SiLU block followed by a zero 
convolution~\cite{controlnet}, whose zero initialization allows the 
pretrained diffusion backbone to remain undisturbed at the start of 
training and gradually learn the optimal embedding. The four refined 
features are fed to the corresponding Condition Modulation.

\subsubsection{\textbf{Condition Modulation}}
To strengthen condition guidance at each resolution, we introduce a
Condition Modulation (CondMod), shown in Fig.~\ref{fig:g_ipr}(b).
At every scale, parallel CondMod fuse the noise feature, the 
condition feature, and the textual embedding.

Following the FiLM-style modulation~\cite{perez2018film, dhariwal2021diffusion},
the noise feature is first normalized via GroupNorm to obtain 
$\mathcal{N}_{norm}$, and the condition feature is passed through two parallel 
depth-wise convolutions to predict the scale and shift parameters 
$\alpha$ and $\beta$:
\begin{equation}
\mathbf{X}_{\mathrm{mid}} = \mathcal{N}_{norm} \odot (1 + \alpha) + \beta,
\label{eq:condmod}
\end{equation}
where $\odot$ denotes element-wise multiplication. The $(1+\alpha)$ 
formulation initializes the block close to an identity mapping, 
stabilizing training. The modulated feature $\mathbf{X}_{\mathrm{mid}}$ 
is then added to the textual embedding to yield the block output, 
achieving spatially-aware conditioning driven by the visual control signal 
and semantically-aware conditioning driven by the prompt.

\subsection{Detection Framework} %
\label{sec:det}
\subsubsection{\textbf{Inverted Residual Shift Attention}}
PCB defects are typically small, low-contrast, and easily masked by 
complex circuit patterns. Self-attention captures global semantics but 
is insensitive to fine textures, while convolutions capture local 
details but lack long-range context; neither alone is sufficient for 
reliable small-defect detection.

\begin{figure}[!h]  
	\centerline{\includegraphics[page=1,trim = 0mm 0mm 0mm 0mm, clip, width=1\linewidth]{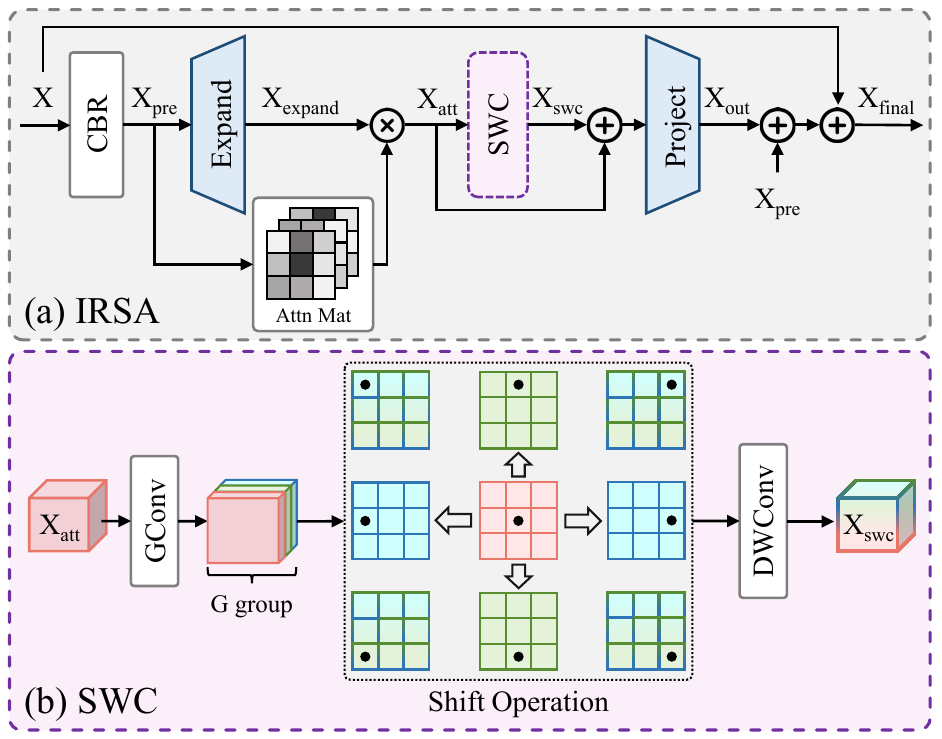}}
  \captionsetup{skip=0pt}
	\caption{(a) Overview of the Inverted Residual Shift Attention Block. (b) Structure of the shift-wise convolution (SWC).}  
	\label{fig:irsa}
	\vspace{-0.5em}
\end{figure}

We therefore propose the Inverted Residual Shift Attention (IRSA) 
block (Fig.~\ref{fig:irsa}(a)), which couples self-attention and 
shift-wise convolution in cascaded within an inverted-residual 
structure. Given input $\mathbf{X}\in\mathbb{R}^{C\times H\times W}$, 
we first apply a convolution–batch normalization–ReLU (CBR) layer to 
obtain a preliminary feature $\mathbf{X}_{\text{pre}} = \mathrm{CBR}(\mathbf{X})$, 
and then expand the channel dimension from $C$ to $C'$ via a 
$1{\times}1$ convolution, yielding $\mathbf{X}_{\text{expand}}$. Two cascaded modules are then employed:
\begin{equation}
\mathbf{X}_{{att}} = {Softmax}\!\left(
\frac{\mathbf{Q}\mathbf{K}^{\!\top}}{\sqrt{d_k}}\right)\mathbf{V}, 
\end{equation}
\begin{equation}
\mathbf{X}_{{swc}} = {DWConv}\!\left(
{Shift}_{G}(\mathbf{X}_{{att}})\right),
\end{equation}
where $\mathbf{Q}$ and $\mathbf{K}$ are linear projections of $\mathbf{X}_{{pre}}$, and $\mathbf{V}$ is a linear projections of 
$\mathbf{X}_{{expand}}$. ${Shift}_G(\cdot)$ splits 
channels into $G$ groups and spatially shifts each group along one 
of $G$ pre-defined directions (up, down, left, right, and four 
diagonals)~\cite{shift_wise}, followed by a depth-wise convolution that 
aggregates the shifted context. The branch outputs are merged and 
projected back to $C$ channels:
\begin{equation}
\mathbf{X}_{{out}} = {Conv}_{1\times 1}(
\mathbf{X}_{{att}} + \mathbf{X}_{{swc}}).
\end{equation}

The final output restores the input and the preliminary feature paths to ensure stable gradient flow:
\begin{equation}
\mathbf{X}_{{final}} = \mathbf{X} + \mathbf{X}_{{pre}} 
+ \mathbf{X}_{{out}}.
\end{equation}

The cascaded coupling of global attention and shift-wise convolution, 
together with the multi-level residuals, yields efficient yet 
discriminative features for small PCB defects.

\subsubsection{\textbf{Cross-level Complementary Fusion}}
\begin{figure}[!h]  
	\centerline{\includegraphics[page=1,trim = 0mm 0mm 0mm 0mm, clip, width=1\linewidth]{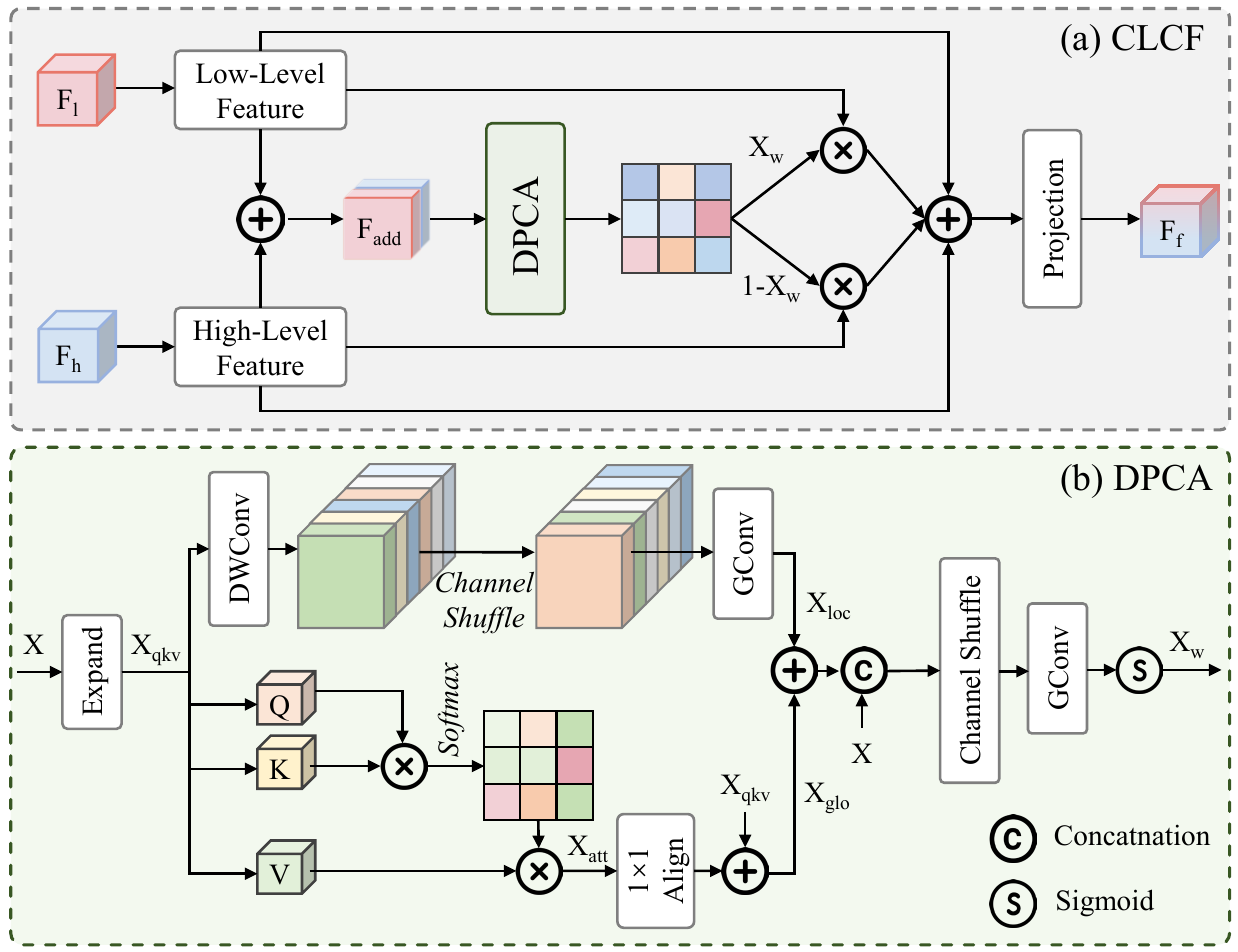}}
  \captionsetup{skip=0pt}
	\caption{(a) Overview of the Cross-level Complementary Fusion (CLCF) Block. (b) Structure of the Dual-path Collaborative Attention (DPCA) Block.}  
	\label{fig:CLCF}
	\vspace{-1em}
\end{figure}

Shallow features retain fine textures crucial for localizing tiny 
defects but are noise-sensitive, whereas deep features carry strong 
semantics but lose spatial detail. Naive addition or concatenation 
tends to amplify noise or suppress subtle defects. To better fuse 
the two, we propose the Cross-level Complementary Fusion (CLCF), comprising a Dual-path Collaborative Attention (DPCA) that 
predicts a pixel-wise gate, and a gated weighting scheme with residual 
compensation (Fig.~\ref{fig:CLCF}(a)).

\emph{DPCA sub-block.} Given the concatenated input 
$\mathbf{X}\in\mathbb{R}^{C\times H\times W}$, a $1{\times}1$ 
convolution produces the shared tensor $\mathbf{X}_{qkv}$. Two parallel 
branches then operate on it. The \emph{local} branch captures 
fine-grained cues:
\begin{equation}
\mathbf{X}_{\text{loc}} = \mathrm{GConv}\!\left(\mathrm{CS}\!\left(
\mathrm{DWConv}_{3\times 3}(\mathbf{X}_{qkv})\right)\right),
\end{equation}
where $\mathrm{CS}(\cdot)$ denotes channel shuffle. The \emph{global} 
branch computes self-attention with a learnable per-head scaling 
vector $\boldsymbol{\alpha}$, following~\cite{ali2021xcit}:
\begin{equation}
\mathbf{X}_{\text{glo}} = \mathrm{Conv}_{1\times 1}\!\left(
\mathrm{Softmax}\!\left(\frac{\mathbf{Q}\mathbf{K}^{\top}}
{\boldsymbol{\alpha}}\right)\mathbf{V}\right) + \mathbf{X}_{qkv}.
\end{equation}
The two outputs are concatenated with the input and passed through a 
grouped convolution with \emph{sigmoid} activation to yield a 
pixel-wise gate:
\begin{equation}
\mathbf{w} = \sigma\!\left(\mathrm{GConv}_{7\times 7}\!\left(
\mathrm{CS}\!\left(\mathrm{Concat}[\mathbf{X}_{\text{loc}},\mathbf{X}_{\text{glo}},\mathbf{X}]\right)\right)\right),
\end{equation}

\emph{Gated fusion.} Let $\mathbf{F}_l$ and $\mathbf{F}_h$ denote the 
low- and high-level inputs aligned to the same resolution. CLCF fuses 
them via a gate-and-residual scheme:
\begin{equation}
\mathbf{F}_f = \mathrm{Proj}\!\left(\mathbf{w}\odot \mathbf{F}_l 
+ (1-\mathbf{w})\odot \mathbf{F}_h + (\mathbf{F}_l + \mathbf{F}_h)\right),
\end{equation}
where the residual term $\mathbf{F}_l+\mathbf{F}_h$ prevents information 
loss from over-gating. The gate $\mathbf{w}$ adaptively selects detail 
cues from $\mathbf{F}_l$ where defects are subtle and semantic cues 
from $\mathbf{F}_h$ elsewhere, yielding features well suited for the 
detection head.

\begin{figure}[!t]  
	\centerline{\includegraphics[page=1,trim = 0mm 0mm 0mm 0mm, clip, width=1\linewidth]{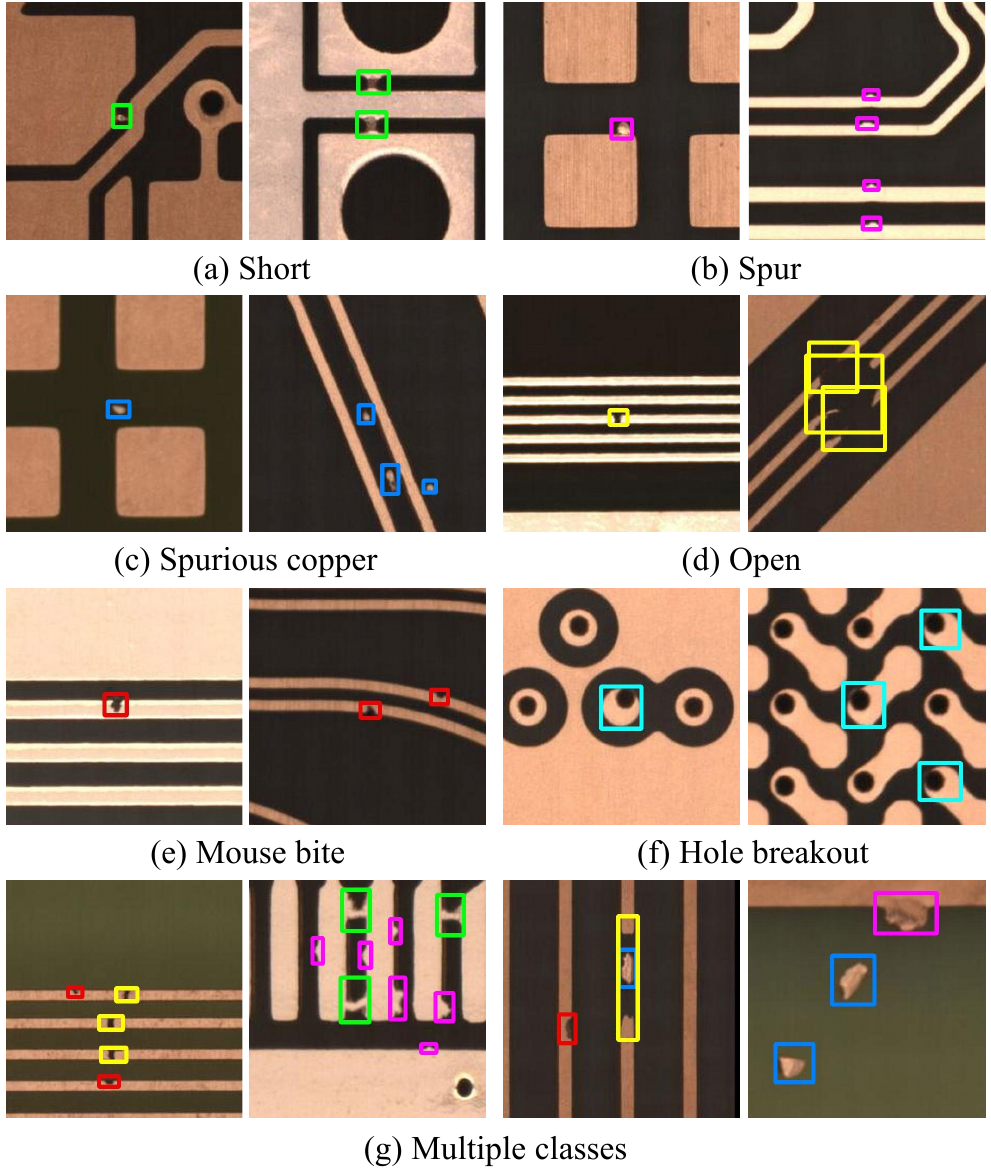}}
  \captionsetup{skip=0pt}
	\caption{Representative annotated samples from the six defect categories, including single-class images (a--f) and a multi-class example (g). (a) Short; (b) Spur; (c) Spurious copper; (d) Open; (e) Mouse bite; (f) Hole breakout; (g) Multiple classes.}  
	\label{fig:data_sample}
	\vspace{-1.5em}
\end{figure}

\section{Experiments}

\subsection{Datasets}

\rev{All reported generation and detection experiments use only the public PCB defect dataset DsPCBSD+~\cite{DsPCBSD+}; no private or newly collected PCB images are included. Six defect categories that significantly affect PCB quality are selected: short, spur, spurious copper, open, mouse bite, and hole breakout~\cite{hripcb,deeppcb}. Annotation errors in the original dataset are corrected before constructing the generation and detection subsets, and representative corrected samples are shown in Fig.~\ref{fig:data_sample}.}

\noindent\textbf{Generation dataset.} A subset of 5{,}252 defect 
images is reconstructed from DsPCBSD+ to train the diffusion model, 
as detailed in Tab.~\ref{tab:data}. The training and validation sets 
are split at a ratio of 8:2.

\noindent\textbf{Detection dataset.} The remaining images are 
reorganized into a base detection set of 2{,}027 images 
(Tab.~\ref{tab:data}, column ``Base''), with all defects manually 
annotated using rectangular bounding boxes and split 8:2 into 
training and validation sets. To investigate the effect of different 
data expansion strategies, two extended variants are constructed:

\begin{itemize}
    \item \textbf{Extend~I} (3{,}319 images): the base set expanded 
    via conventional augmentations, including random flipping, 
    rotation, and Gaussian blur.
    \item \textbf{Extend~II} (3{,}382 images): the base set expanded 
    by incorporating defect images synthesized by the trained 
    diffusion model. Annotations are produced semi-automatically: 
    a seed set of real images is manually labeled, an auxiliary 
    annotation model is trained on this seed set to label the 
    synthetic images, and a final manual verification step corrects 
    residual errors.
\end{itemize}

The two extended datasets are used in Sec.~\ref{sec:result} to 
provide a controlled comparison between traditional and 
generation-based augmentation strategies.

\subsection{Experimental Setup} 
\subsubsection{\textbf{Implementation Details}} 
The experimental environment in this study runs on an Ubuntu system, and the experimental setting employs an Intel Xeon E5-2686 v4 CPU and four 24G NVIDIA TITAN RTX GPUs as the hardware configuration. The generation model is based on Stable Diffusion v1.5 and uses a batch size of 1, 30,000 training steps, and a learning rate of 0.00002, while the detection model adopts a batch size of 4, is trained for 100 epochs, and uses a learning rate of 0.0001. For both the generation and detection tasks, the input image resolution is set to 512×512. All networks are optimized using the AdamW optimizer, where the momentum and weight decay are set to 0.9 and 0.0001, respectively.

\subsubsection{\textbf{Evaluation Metrics}} 

The quality of synthesized PCB defect images is evaluated from 
three complementary perspectives. Fréchet Inception Distance 
(FID)~\cite{fid} measures the distributional similarity between 
real and generated images in the feature space of a pretrained 
Inception network, where lower values indicate closer distributions:
\begin{equation}
\mathrm{FID} = \|\mu_r - \mu_g\|^2 +
\mathrm{Tr}\!\left(\Sigma_r + \Sigma_g -
2\left(\Sigma_r\Sigma_g\right)^{\frac{1}{2}}\right),
\end{equation}
where $\mu_r, \Sigma_r$ and $\mu_g, \Sigma_g$ are the mean and 
covariance of the real and generated feature distributions, 
respectively. Learned Perceptual Image Patch Similarity 
(LPIPS)~\cite{lpips} quantifies perceptual similarity by comparing 
deep feature activations across $L$ layers weighted by $w_l$, 
with lower values indicating greater perceptual fidelity:
\begin{equation}
\mathrm{LPIPS} = \sum_{l} w_l \cdot
\|\phi_l(x) - \phi_l(y)\|_2^2.
\end{equation}
Peak Signal-to-Noise Ratio (PSNR) and Structural Similarity Index 
Measure (SSIM) \cite{ssim} assess pixel-level reconstruction quality and 
structural consistency, respectively, with higher values indicating 
better fidelity:
\begin{equation}
\mathrm{PSNR} = 10 \cdot \log_{10}\!\left(
\frac{\mathrm{MAX}^2}{\mathrm{MSE}}\right),
\end{equation}
\begin{equation}
\mathrm{SSIM} = \frac{(2\mu_x\mu_y + C_1)(2\sigma_{xy} + C_2)}
{(\mu_x^2 + \mu_y^2 + C_1)(\sigma_x^2 + \sigma_y^2 + C_2)},
\end{equation}
where $\mu_x, \mu_y$ are local means, $\sigma_x^2, \sigma_y^2$ are 
variances, $\sigma_{xy}$ is the cross-covariance, and $C_1, C_2$ 
are stabilizing constants.

\noindent\textbf{Detection metrics.}
The downstream defect detection performance is evaluated using 
Precision, Recall, and mean Average Precision at two IoU thresholds 
(mAP@0.5 and mAP@0.5:0.95). Precision measures the fraction of 
correct detections among all detections, Recall measures the 
fraction of ground-truth defects successfully detected, and mAP 
summarizes the area under the precision--recall curve averaged over 
$N$ defect categories:
\begin{equation}
\mathrm{Precision} = \frac{\mathrm{TP}}{\mathrm{TP} + \mathrm{FP}},
\quad
\mathrm{Recall} = \frac{\mathrm{TP}}{\mathrm{TP} + \mathrm{FN}},
\end{equation}
\begin{equation}
\mathrm{mAP} = \frac{1}{N}\sum_{c=1}^{N}
\int_0^1 P(R)\,\mathrm{d}R.
\end{equation}

\begin{table}[!tbp]
\setlength{\abovecaptionskip}{2pt}
\caption{Dataset statistics for defect generation and detection 
         experiments. Extend~\textbf{I}: traditional augmentation; 
         Extend~\textbf{II}: generation-based augmentation.
         \label{tab:data}}
\centering
\renewcommand\arraystretch{1.15}
\resizebox{\linewidth}{!}{
\begin{tabular}{c|c|c|ccc}

\toprule[1pt]
\toprule[0.5pt]

\multirow{2}{*}{Type} & \multirow{2}{*}{Class} 
& \multirow{2}{*}{\makecell{Defect\\Gen.}} 
& \multicolumn{3}{c}{Defect Det.} \\
\cline{4-6}
& & & Base & Extend \textbf{I} & Extend \textbf{II} \\
\midrule
\multirow{7}{*}{\rotatebox{90}{\small No.~of Images}}
& Short          & 180   & 137   & 217   & 579 \\
& Spur           & 1,326 & 444   & 705   & 597 \\
& Spurious copper& 826   & 360   & 597   & 613 \\
& Open           & 770   & 346   & 564   & 768 \\
& Mouse bite     & 1,084 & 422   & 705   & 592 \\
& Hole breakout  & 1,066 & 433   & 715   & 540 \\
& \textbf{Total} & \textbf{5,252} & \textbf{2,027} 
                 & \textbf{3,319} & \textbf{3,382} \\
\midrule
\multirow{7}{*}{\rotatebox{90}{\small No.~of Defects}}
& Short          & 228   & 161   & 253   & 839 \\
& Spur           & 2,213 & 546   & 874   & 747 \\
& Spurious copper& 969   & 385   & 641   & 667 \\
& Open           & 919   & 362   & 594   & 936 \\
& Mouse bite     & 1,309 & 440   & 732   & 636 \\
& Hole breakout  & 2,342 & 480   & 796   & 615 \\
& \textbf{Total} & \textbf{7,980} & \textbf{2,374} 
                 & \textbf{3,890} & \textbf{4,440} \\
\bottomrule[1pt]
\end{tabular}}
\vspace{-1em}
\end{table}

\subsection{Comparative Experiment} 
\label{sec:result}

\subsubsection{\textbf{Detection Performance Comparison}} 
To comprehensively evaluate UniPCB, we compare it against ten 
representative detectors spanning two architectural families. 
The Transformer-based group includes DETR~\cite{DETR}, 
Deformable-DETR~\cite{deformable-DETR}, DAB-DETR~\cite{dab-DETR}, 
DINO~\cite{dino}, RT-DETR~\cite{RT-DETR}, D-FINE~\cite{D-Fine}, 
and DEIM~\cite{deim}; the CNN-based group includes 
YOLOv8~\cite{yolov8}, YOLOv10~\cite{yolov10}, and 
YOLO11~\cite{yolo11}. RT-DETR serves as the direct baseline, 
as UniPCB is built upon it.

\begin{figure*}[!t]  
	\centerline{\includegraphics[page=1,trim = 0mm 0mm 0mm 0mm, clip, width=0.8\linewidth]{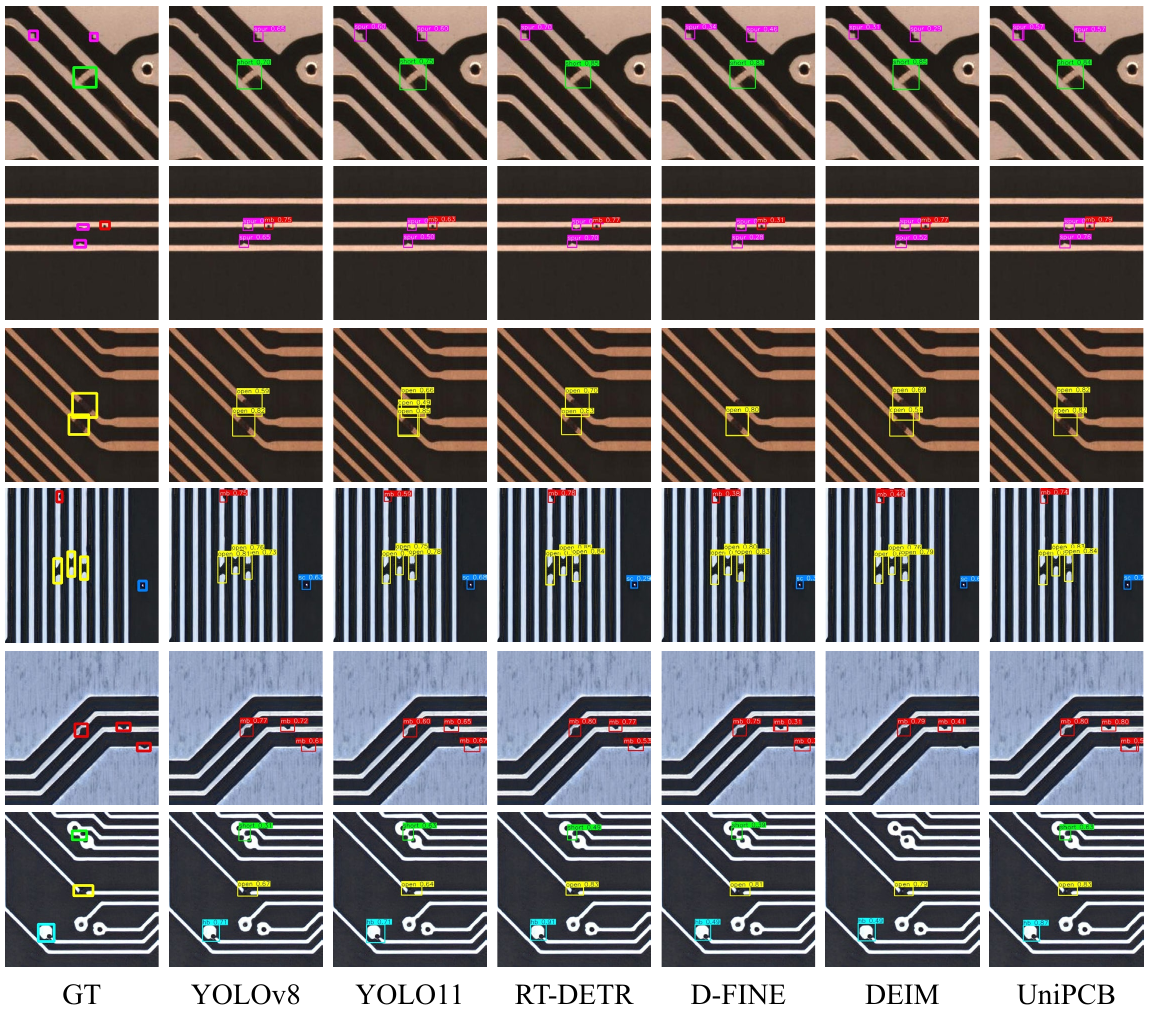}}
  \captionsetup{skip=0pt}
	\caption{Visualization of detection results across different models. Zoom-in for best view.}  
	\label{fig:model_cmp}
	\vspace{-1.5em}
\end{figure*}

Quantitative results are reported in Tab.~\ref{tab:detec_cmp}, 
and visual comparisons are shown in Fig.~\ref{fig:model_cmp}. 
UniPCB achieves Precision / Recall / mAP@0.5 / mAP@0.5:0.95 of 
0.977 / 0.953 / 0.980 / 0.618, outperforming all compared methods. 
Relative to the RT-DETR baseline, mAP@0.5 improves by \rev{2.1 percentage points} and 
mAP@0.5:0.95 by \rev{1.7 percentage points}, demonstrating that the IRSA and CLCF modules 
provide consistent gains beyond the base architecture. Among 
Transformer-based methods, the strongest competitor is DEIM 
(mAP@0.5: 0.958), which UniPCB surpasses by \rev{2.2 percentage points}. Among CNN-based 
methods, YOLOv8 achieves the best mAP@0.5:0.95 of 0.611, still 
\rev{0.7 percentage points} below UniPCB. The visual results in Fig.~\ref{fig:model_cmp} 
further \rev{illustrate} that UniPCB produces fewer missed detections and 
tighter bounding boxes on small, low-contrast defects, where most 
competing methods show visible localization errors or omissions. \rev{Because all detectors in this comparison use the same Extend~II data, the results isolate the contribution of the detection architecture from differences in training data. Together with the module ablation, these results show that IRSA and CLCF provide architecture-side gains under generation-augmented training.}

\begin{table*}[!tbp] 
\setlength{\abovecaptionskip}{2pt}
  \caption{Detection Performance of Different Models. The Top Three Results are Marked with \textcolor{red}{{red}}, \textcolor{blue}{{blue}}, and \textcolor{green}{{green}}, respectively. \label{tab:detec_cmp}}
  \centering
  \renewcommand\arraystretch{1.1}	
    \resizebox{0.9\linewidth}{!}{\begin{tabular}{l|c|cccc|ccc} 
  
    \toprule[1pt]
    \toprule[0.5pt]

    \rowcolor[HTML]{EDEDED}
    Model
    & Source
    &  Precision$\uparrow$
    &  Recall$\uparrow$
    &  mAP@0.5$\uparrow$
    &  mAP@0.5:0.95$\uparrow$
    &  FPS$\uparrow$
    &  FLOPS(G)$\downarrow$
    &  Param.(M)$\downarrow$
    \\

    \midrule
    DETR \cite{DETR}
    & 20‘ECCV
    & 0.896 & 0.875 & 0.923 & 0.582
    & 66.6 & 60.5 & 41.5
    \\

    Deformable-DETR \cite{deformable-DETR}
    & 21'ICLR
    & 0.915 & 0.905 & 0.940 & 0.592
    & 32.5 & 125.9 & 40.1
    \\

    DAB-DETR \cite{dab-DETR}
    & 22'ICLR
    & 0.872 & 0.889 & 0.905 & 0.570
    & 41.0 & 65.3 & 43.7
    \\

    DINO \cite{dino}
    & 23’ICLR
    & 0.922 & 0.913 & 0.948 & 0.598
    & 23.5 & 183.3 & 47.5
    \\
    
    YOLOv8 \cite{yolov8}
    & 23'Ultralytics
    & \textcolor{green}{0.954} & \textcolor{blue}{0.941} & \textcolor{blue}{0.965} & \textcolor{blue}{0.611}
    & 192.8 & 23.4 & 9.83
    \\

    YOLOv10 \cite{yolov10}
    & 24'NIPS
    & 0.850 & 0.855 & 0.911 & 0.583 
    & 168.5 & 24.5 & 8.04
    \\

    RT-DETR \cite{RT-DETR}
    & 24'CVPR
    & 0.949 & 0.921 & 0.959 & 0.601 
    & 97.0 & 57.2 & 19.9 
    \\

    YOLO11 \cite{yolo11}
    & 24'Ultralytics
    & \textcolor{blue}{0.956} & 0.921 & \textcolor{green}{0.960} & 0.607
    & 192.3 & 21.3 & 9.42
    \\

    D-FINE \cite{D-Fine}
    & 25'ICLR
    & 0.938 & 0.927 & 0.952 & 0.606
    & 150.6 & 56.3 & 19.1
    \\

    DEIM \cite{deim}
    & 25'CVPR
    & 0.942 & \textcolor{green}{0.938} & 0.958 & \textcolor{green}{0.610}
    & 118.2 & 92.5 & 31.3
    \\
    
    \rowcolor[HTML]{FCF0FF}
    UniPCB (Detection)
    & -
    & \textcolor{red}{0.977} & \textcolor{red}{0.953} & \textcolor{red}{0.980} & \textcolor{red}{0.618}
    & 90.9 & 71.8 & 19.1
    \\

  \bottomrule[1pt]
  \end{tabular}}
  \vspace{-1em}
\end{table*}

\subsubsection{\textbf{Generation Performance Comparison}}

\begin{figure}[!htbp]  
	\centerline{\includegraphics[page=1,trim = 0mm 0mm 0mm 0mm, clip, width=1\linewidth]{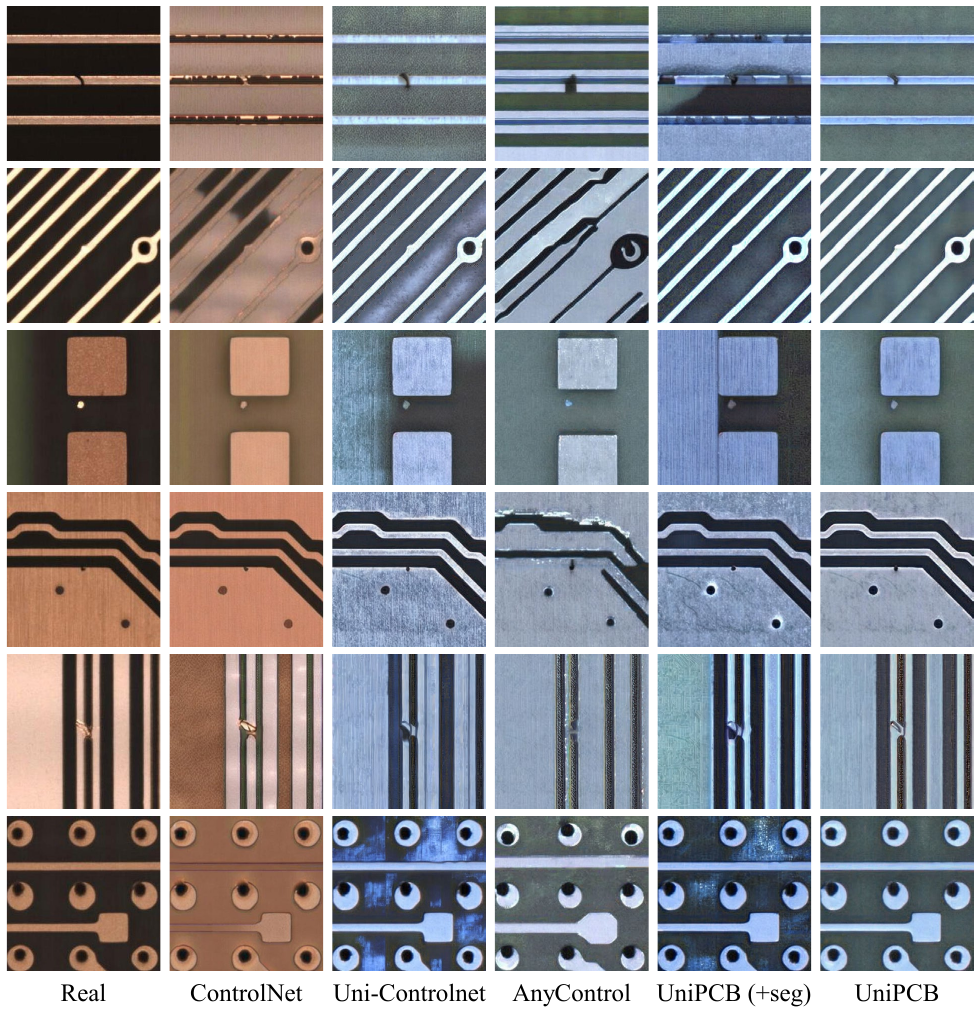}}
  \captionsetup{skip=0pt}
	\caption{Visualization of generation results across different models. For better clarity, results are best viewed in enlarged form.}  
	\label{fig:gen_cmp}
	\vspace{-1em}
\end{figure}

We compare UniPCB against three representative conditional generation 
baselines: ControlNet~\cite{controlnet}, which injects a single 
condition at the initial resolution; Uni-ControlNet~\cite{uni-controlnet}, 
which combines edge and depth conditions through a unified control 
mechanism but still injects them at a single scale; and 
AnyControl~\cite{anycontrol}, which supports flexible multi-condition 
combinations but without explicit cross-modal fusion. To further 
isolate the effect of condition type, we also evaluate a variant 
UniPCB~(+seg) that replaces the text condition with a defect 
segmentation mask as the third control signal. Results are reported 
in Tab.~\ref{tab:gen_cmp} and visualized in Fig.~\ref{fig:gen_cmp}.

The results reveal a clear progression. ControlNet, relying solely 
on edge maps, achieves PSNR/SSIM of 10.23/0.502 but yields a 
relatively high FID of 131.11, as single-modality conditions cannot 
fully constrain the structural diversity of PCB layouts. 
Uni-ControlNet improves PSNR to 10.60 by adding depth as a 
complementary condition, yet its FID rises to 145.53 and SSIM drops 
to 0.457, suggesting that naïve concatenation of conditions without 
dedicated fusion introduces inter-modal interference. AnyControl 
shows similar limitations despite its flexible condition 
combinations, achieving the weakest overall performance 
(FID: 142.01, SSIM: 0.404). UniPCB~(+seg) improves PSNR and SSIM 
over the baselines by adding a segmentation mask, but its FID 
(142.93) and LPIPS (0.535) remain worse than UniPCB, indicating 
that a pixel-level mask introduces spatial redundancy with the edge 
condition rather than complementary information.

UniPCB achieves the best performance across all metrics 
(FID: 129.61, LPIPS: 0.457, PSNR: 13.45, SSIM: 0.619). The key 
distinction from prior methods is not the number of conditions but 
their complementarity and fusion quality: edge, depth, and text 
provide structural, spatial, and semantic cues that cover orthogonal 
axes of control, while ScaleEncoder and CondMod ensure that these 
conditions are effectively embedded across multiple scales rather 
than injected at a single resolution. The qualitative results in 
Fig.~\ref{fig:gen_cmp} further confirm that UniPCB preserves the 
global circuit topology while rendering localized defect patterns 
with higher fidelity than all compared methods.

\begin{table}[!htbp]
\setlength{\abovecaptionskip}{0pt}
  \caption{Generation Performance of Different Models. \label{tab:gen_cmp}}
  \centering
  \renewcommand\arraystretch{1.2}
  {\begin{tabular}{l|cccc}
    \toprule[1pt]
    \rowcolor[HTML]{EDEDED}
    Model & FID$\downarrow$ & LPIPS$\downarrow$ & PSNR$\uparrow$ & SSIM$\uparrow$ \\
    \midrule
    ControlNet \cite{controlnet} & 131.11 & 0.566 & 10.23 & 0.502 \\
    Uni-Controlnet \cite{uni-controlnet} & 145.53 & 0.562 & 10.60 & 0.457 \\
    AnyControl \cite{anycontrol} & 142.01 & 0.635 & 9.02 & 0.404 \\
    UniPCB (+seg) & 142.93 & 0.535 & 12.47 & 0.525 \\
    \rowcolor[HTML]{FCF0FF}
    UniPCB (ours) & \textbf{129.61} & \textbf{0.457} & \textbf{13.45} & \textbf{0.619} \\
    \bottomrule[1pt]
  \end{tabular}}
  \vspace{-1em}
\end{table}

\begin{table*}[!tbp] 
\setlength{\abovecaptionskip}{0pt}
  \caption{
  Evaluation of augmentation strategies. Superscripts $\mathrm{I}$ and $\mathrm{II}$ denote traditional augmentation and generative augmentation, respectively. For results obtained with generative augmentation, the percentages in parentheses indicate relative changes compared with the corresponding traditional augmentation setting. \textcolor{red}{red} indicates favorable changes, while \textcolor{green}{green} indicates unfavorable changes.
  \label{tab:aug_cmp}}
  \centering
  \renewcommand\arraystretch{1.2}	
    {\begin{tabular}{l|c|cccc} 
  
    \toprule[1pt]
    \toprule[0.5pt]

    \rowcolor[HTML]{EDEDED}
    Model
    & Source
    &  Precision$\uparrow$
    &  Recall$\uparrow$
    &  mAP@0.5$\uparrow$
    &  mAP@0.5:0.95$\uparrow$
    \\

    \midrule
    DETR\textsuperscript{I} \cite{DETR}
    & \multirow{2}{*}{\makecell{20'ECCV}}
    & 0.892 & 0.848 & 0.914 & 0.556
    \\
    DETR\textsuperscript{II} \cite{DETR}
    & 
    & 0.896 $\textcolor{red}{\,\uparrow0.4\%}$ & 0.875 $\textcolor{red}{\,\uparrow2.7\%}$ & 0.923 $\textcolor{red}{\,\uparrow0.9\%}$ & 0.582 $\textcolor{red}{\,\uparrow2.6\%}$
    \\

    \hline
    Deformable-DETR\textsuperscript{I} \cite{deformable-DETR}
    & \multirow{2}{*}{\makecell{21'ICLR}}
    & 0.917 & 0.882 & 0.932 & 0.561
    \\
    Deformable-DETR\textsuperscript{II} \cite{deformable-DETR}
    & 
    & 0.915 $\textcolor{green}{\,\downarrow 0.2\%}$ & 0.905 $\textcolor{red}{\,\uparrow2.3\%}$ & 0.940 $\textcolor{red}{\,\uparrow0.8\%}$ & 0.592 $\textcolor{red}{\,\uparrow3.1\%}$
    \\

    \hline
    DAB-DETR\textsuperscript{I} \cite{dab-DETR}
    & \multirow{2}{*}{\makecell{22'ICLR}}
    & 0.869 & 0.867 & 0.899 & 0.545
    \\
    DAB-DETR\textsuperscript{II} \cite{dab-DETR}
    & 
    & 0.872 $\textcolor{red}{\,\uparrow0.3\%}$ & 0.889 $\textcolor{red}{\,\uparrow2.2\%}$ & 0.905 $\textcolor{red}{\,\uparrow0.6\%}$ & 0.570 $\textcolor{red}{\,\uparrow2.5\%}$
    \\

    \hline
    DINO\textsuperscript{I} \cite{dino}
    & \multirow{2}{*}{\makecell{23’ICLR}}
    & 0.928 & 0.888 & 0.940 & 0.574
    \\
    DINO\textsuperscript{II} \cite{dino}
    & 
    & 0.922 $\textcolor{green}{\,\downarrow 0.6\%}$ & 0.913 $\textcolor{red}{\,\uparrow2.5\%}$ & 0.948 $\textcolor{red}{\,\uparrow0.8\%}$ & 0.598 $\textcolor{red}{\,\uparrow2.4\%}$
    \\
    
    \hline
    YOLOv8\textsuperscript{I} \cite{yolov8}
    & \multirow{2}{*}{\makecell{23'Ultralytics}}
    & 0.945 & 0.933 & 0.956 & 0.601
    \\
    YOLOv8\textsuperscript{II} \cite{yolov8}
    & 
    & 0.954 $\textcolor{red}{\,\uparrow0.9\%}$ & 0.941 $\textcolor{red}{\,\uparrow0.8\%}$ & 0.965 $\textcolor{red}{\,\uparrow0.9\%}$ & 0.611 $\textcolor{red}{\,\uparrow1.0\%}$
    \\

    \hline
    YOLOv10\textsuperscript{I} \cite{yolov10}
    & \multirow{2}{*}{\makecell{24'NIPS}}
    & 0.853 & 0.840 & 0.901 & 0.564 
    \\
    YOLOv10\textsuperscript{II} \cite{yolov10}
    & 
    & 0.850 $\textcolor{green}{\,\downarrow 0.3\%}$ & 0.855 $\textcolor{red}{\,\uparrow1.5\%}$ & 0.911 $\textcolor{red}{\,\uparrow1.0\%}$ & 0.583 $\textcolor{red}{\,\uparrow1.9\%}$
    \\

    \hline
    RT-DETR\textsuperscript{I} \cite{RT-DETR}
    & \multirow{2}{*}{\makecell{24'CVPR}}
    & 0.951 & 0.895 & 0.950 & 0.573 
    \\
    RT-DETR\textsuperscript{II} \cite{RT-DETR}
    & 
    & 0.949 $\textcolor{green}{\,\downarrow 0.3\%}$ & 0.921 $\textcolor{red}{\,\uparrow2.6\%}$ & 0.959 $\textcolor{red}{\,\uparrow0.9\%}$ & 0.601 $\textcolor{red}{\,\uparrow2.8\%}$ 
    \\

    \hline
    YOLO11\textsuperscript{I} \cite{yolo11}
    & \multirow{2}{*}{\makecell{24'Ultralytics}}
    & 0.927 & 0.899 & 0.952 & 0.574
    \\
    YOLO11\textsuperscript{II} \cite{yolo11}
    & 
    & 0.956 $\textcolor{red}{\,\uparrow2.9\%}$ & 0.921 $\textcolor{red}{\,\uparrow2.2\%}$ & 0.960 $\textcolor{red}{\,\uparrow0.8\%}$ & 0.607 $\textcolor{red}{\,\uparrow3.3\%}$
    \\

    \hline
    D-FINE\textsuperscript{I} \cite{D-Fine}
    & \multirow{2}{*}{\makecell{25'ICLR}}
    & 0.935 & 0.910 & 0.944 & 0.579
    \\
    D-FINE\textsuperscript{II} \cite{D-Fine}
    & 
    & 0.938 $\textcolor{red}{\,\uparrow0.3\%}$ & 0.927 $\textcolor{red}{\,\uparrow1.7\%}$ & 0.952 $\textcolor{red}{\,\uparrow0.8\%}$ & 0.606 $\textcolor{red}{\,\uparrow2.7\%}$
    \\

    \hline
    DEIM\textsuperscript{I} \cite{deim}
    & \multirow{2}{*}{\makecell{25'CVPR}}
    & 0.940 & 0.914 & 0.948 & 0.584
    \\
    DEIM\textsuperscript{II} \cite{deim}
    & 
    & 0.942 $\textcolor{red}{\,\uparrow0.3\%}$ & 0.938 $\textcolor{red}{\,\uparrow2.4\%}$ & 0.958 $\textcolor{red}{\,\uparrow1.0\%}$ & 0.610 $\textcolor{red}{\,\uparrow2.6\%}$
    \\

    \hline
    \rowcolor[HTML]{FCF0FF}
    UniPCB\textsuperscript{I}
    & \multirow{2}{*}{\makecell{-}}
    & 0.970 & 0.927 & 0.971 & 0.593
    \\
    
    \rowcolor[HTML]{FCF0FF}
    UniPCB\textsuperscript{II}
    & 
    & 0.977 $\textcolor{red}{\,\uparrow0.7\%}$ & 0.953 $\textcolor{red}{\,\uparrow2.6\%}$ & 0.980 $\textcolor{red}{\,\uparrow0.9\%}$ & 0.618 $\textcolor{red}{\,\uparrow2.1\%}$
    \\

  \bottomrule[1pt]
  \end{tabular}}
  \vspace{-1em}
\end{table*}

\begin{table*}[!htb] 
\setlength{\abovecaptionskip}{2pt}
  \caption{Ablation Study on The Effectiveness of IRSA and CLCF Modules.\label{tab:abla_exp}}
  \centering
  \renewcommand\arraystretch{1.2}	
    {\begin{tabular}{c|cc|cccc} 

    \rowcolor[HTML]{EDEDED}
    \toprule[1pt]
  
    Index
    & IRSA
    & CLCF
    &  Precision$\uparrow$
    &  Recall$\uparrow$
    &  mAP@0.5$\uparrow$
    &  mAP@0.5:0.95$\uparrow$
    \\

    \hline
    Baseline
    &\bf{\ding{55}}
    &\bf{\ding{55}} 
    & 0.949 & 0.921 & 0.959 & 0.601  
    \\

    \cdashline{1-7}
    a
    &\bf{\ding{51}}
    &\bf{\ding{55}} 
    & 0.951 & 0.933 & 0.968 & 0.614 
    \\

    b
    &\bf{\ding{55}} 
    &\bf{\ding{51}} 
    & 0.969 & 0.949 & 0.969 & 0.614 
    \\

    c
    &\bf{\ding{51}}
    &\bf{\ding{51}}
    & \bf{0.977} & \bf{0.953} & \bf{0.980} & \bf{0.618}
    \\

  \bottomrule[0.8pt]
  \end{tabular}}
  \vspace{-1.5em}
\end{table*}


\begin{table}[!htb] 
\setlength{\abovecaptionskip}{2pt}
  \caption{Ablation results of different attention mechanisms in CLCF Block. \label{tab:abla_DPCA}}
  \centering
  \renewcommand\arraystretch{1.2}	
    \resizebox{\linewidth}{!}{\begin{tabular}{l|cccc} 

    \rowcolor[HTML]{EDEDED}
    \toprule[1pt]

    Method
    &  Precision$\uparrow$
    &  Recall$\uparrow$
    &  mAP@0.5$\uparrow$
    &  mAP@0.5:0.95$\uparrow$
    \\

    \hline
    Baseline 
    & 0.949 & 0.921 & 0.959 & 0.601
    \\

    \cdashline{1-5}
    +SE \cite{se}
    & 0.945 & 0.932 & 0.960 & 0.604
    \\

    +CA \cite{ca}
    & 0.936 & 0.908 & 0.945 & 0.582
    \\

    +SimAM \cite{simam}
    & 0.958 & 0.918 & 0.965 & 0.607
    \\

    +EMA \cite{ema}
    & 0.963 & 0.913 & 0.964 & 0.609
    \\
    
    +DPCA (ours)
    & \textbf{0.969} & \textbf{0.949} & \textbf{0.969} & \textbf{0.614}
    \\

  \bottomrule[1pt]
  \end{tabular}}
  \vspace{-1em}
\end{table}

\subsubsection{\textbf{Data Augmentation Comparison}}

To evaluate the downstream benefit of generation-based augmentation, 
all detection models in Tab.~\ref{tab:detec_cmp} are retrained on 
Extend~I and Extend~II respectively, and their performance is 
systematically compared in Tab.~\ref{tab:aug_cmp}.

Models trained on Extend~II consistently outperform those trained on 
Extend~I across almost all metrics. The gains are most pronounced in 
Recall and mAP@0.5:0.95, where most models improve by 1.5\%--2.0\%, 
reflecting that generation-augmented data helps reduce missed 
detections and improves localization quality at stricter IoU 
thresholds. mAP@0.5 also shows consistent improvement close to 1\%. 
Precision remains nearly unchanged or shows slight declines for some 
models; the minor degradation is likely attributable to additional 
false positives caused by enhanced sensitivity to defect features 
introduced by the richer training distribution.

These results confirm that the generation-based augmentation provides 
richer and more informative training samples than conventional 
augmentation, enabling detection models to better capture 
discriminative features across all six defect categories. Notably, 
UniPCB itself benefits from Extend~II with improvements of 
$+$0.7\% / $+$2.6\% / $+$0.9\% / $+$2.1\% in Precision / Recall / 
mAP@0.5 / mAP@0.5:0.95, demonstrating that the generation and 
detection components of UniPCB are mutually reinforcing.

\begin{figure}[!t]  
	\centerline{\includegraphics[page=1,trim = 0mm 0mm 0mm 0mm, clip, width=1\linewidth]{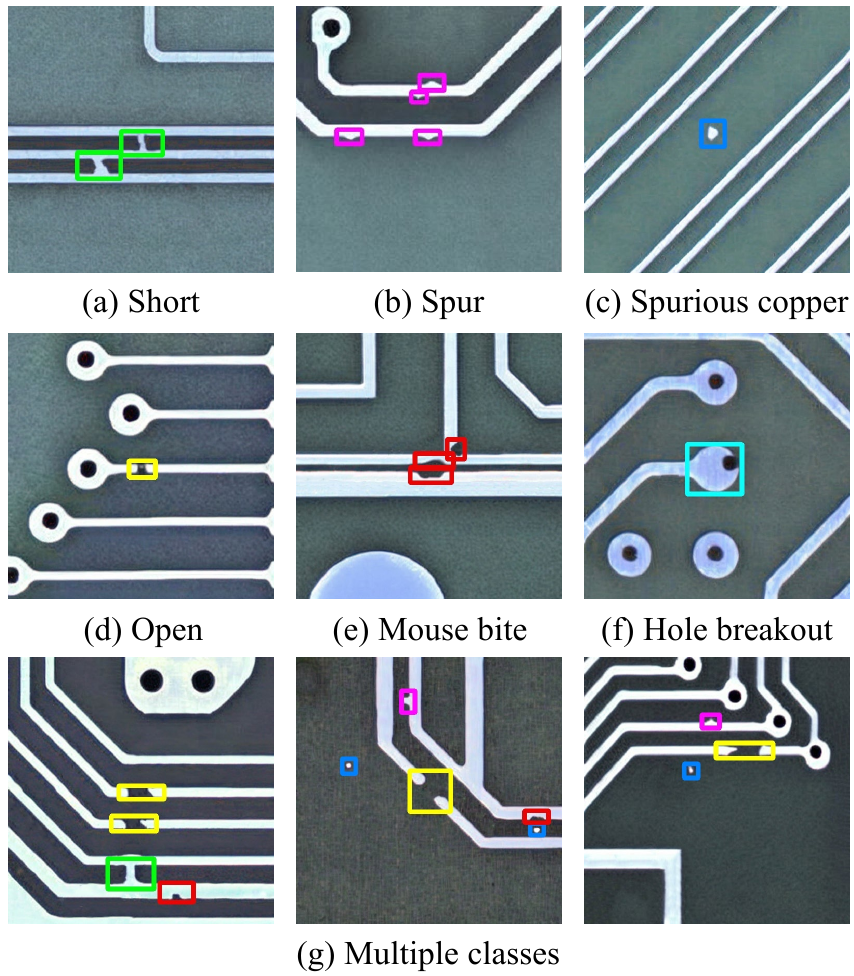}}
  \captionsetup{skip=0pt}
	\caption{Examples of synthesized defect images with bounding-box annotations, covering all six categories in Fig.~\ref{fig:data_sample}. Annotations are generated by an auxiliary model trained on seed annotations and verified manually.}  
	\label{fig:gen_sample}
	\vspace{-1em}
\end{figure}

\subsection{Ablation Study}

\subsubsection{\textbf{Module contribution}}
Tab.~\ref{tab:abla_exp} reports the contribution of each proposed 
module. Adding IRSA alone (setting~(a)) yields gains of 
$+$0.9\%~/~$+$1.3\% in mAP@0.5~/~mAP@0.5:0.95 over the baseline, 
with a notable Recall improvement of $+$1.2\%. This suggests that 
coupling self-attention with shift-wise convolution reduces missed 
detections by capturing fine-grained local cues that vanilla 
self-attention overlooks, particularly for small, low-contrast 
defects embedded in complex circuit patterns. Adding CLCF alone 
(setting~(b)) produces larger Precision gains ($+$2.0\%) alongside 
comparable mAP improvements, indicating that the pixel-level gating 
mechanism effectively suppresses false detections caused by 
background interference during feature fusion. When both modules 
are combined (setting~(c)), the model achieves the best performance 
across all metrics, with mAP@0.5 and mAP@0.5:0.95 reaching 0.980 
and 0.618, improving the baseline by $+$2.1\% and $+$1.7\% 
respectively. The complementary nature of the two modules is 
evident: IRSA strengthens single-scale feature discrimination while 
CLCF enhances cross-level feature aggregation, and their combination 
yields gains exceeding either module alone.

\subsubsection{\textbf{Attention mechanism comparison}}
To validate the design of the DPCA sub-block within CLCF, we replace 
it with four commonly used attention mechanisms and report results in 
Tab.~\ref{tab:abla_DPCA}.

\rev{SE increases Recall by 1.1 percentage points through inter-channel reweighting but provides only limited overall gains. CA reduces mAP@0.5:0.95 by 1.9 percentage points relative to the baseline, whereas SimAM and EMA improve that metric by 0.6 and 0.8 percentage points, respectively. DPCA gives the strongest overall result, reaching Precision / Recall / mAP@0.5 / mAP@0.5:0.95 of 0.969 / 0.949 / 0.969 / 0.614. Its grouped-convolution branch preserves fine spatial cues, while its attention branch models longer-range dependencies; their combination produces spatially adaptive weights for defects at different scales and locations.}

\FloatBarrier
\section{Conclusion and Limitation}

\rev{This paper presents UniPCB, a generation-assisted vision-based measurement framework for the computational processing stage of PCB AOI. The generation branch derives edge, relative pseudo-depth, and text conditions from public PCB images and injects them into a latent diffusion model through ScaleEncoder and CondMod. The detection branch combines IRSA for local--global feature modeling with CLCF for selective cross-level fusion. Experiments on the corrected DsPCBSD+ annotations show improvements in synthetic-image quality and downstream defect classification and localization over the compared baselines. UniPCB therefore contributes to the analysis stage of an AOI instrument; it does not modify the optical sensor or image-acquisition hardware.}

\rev{The generator and detector are trained sequentially rather than optimized end to end. Moreover, all reported results are derived from the public DsPCBSD+ dataset and do not constitute validation across private production lines or AOI devices. The relative pseudo-depth condition is a learned image prior rather than a physical height measurement. Future work will evaluate detector-feedback generation and robustness across board layouts, devices, and illumination.}

\bibliographystyle{IEEEtran}
\bibliography{IEEEabrv,UniPCB}

\end{document}